\documentclass[a4paper, 11pt, onecolumn]{ieeeconf}      

\usepackage{graphicx} 
\usepackage{multirow}
\usepackage{url}
\usepackage[utf8]{inputenc}

\title{\LARGE \bf
Generation of Paths in a Maze using a Deep Network without Learning 
}

\author{Tomas Kulvicius$^{1,*}$, Sebastian Herzog$^{1}$, Minija Tamosiunaite $^{1,2}$ and Florentin W\"org\"otter$^{1}$
\thanks{The research leading to these results has received funding from the  European Community's H2020 Programme (Future and Emerging Technologies,
FET) under grant agreement no. 732266, Plan4Act.}
\thanks{$^{1}$T. Kulvicius, S. Herzog, M. Tamosiunaite and F. W\"org\"otter are with Department for Computational Neuroscience,        University of G\"ottingen, 37073 G\"ottingen, Germany}%
\thanks{$^{2}$M. Tamosiunaite is also with the Faculty of Computer Science, Vytautas Mangnus University, Kaunas, Lithuania}%
\thanks{T.K. contributed to the conception and design of the work, development and implementation of the algorithm, data acquisition, analysis and interpretation of the data, and writing of the paper. S.H. contributed to the implementation of the algorithm, data acquisition, analysis and interpretation of the data. M.T. contributed to the analysis and interpretation of the data, and writing of the paper. F.W. contributed to the conception and design of the work, analysis and interpretation of the data, and writing of the paper.}%
\thanks{$^{*}$Correspondence should be sent to T.K. ({\tt\small tomas.kulvicius@uni-goettingen.de})}%
}

\begin{document}

\maketitle


  {\bf Trajectory- or path-planning is a fundamental issue in a wide variety of applications. Here we show that it is possible to solve path planning for multiple start- and end-points highly efficiently with a network that consists only of max pooling layers, for which no network training is needed. Different from competing approaches, very large mazes containing more than half a billion nodes with dense obstacle configuration and several thousand path end-points can this way be solved in very short time on parallel hardware.}

\section*{\flushleft INTRODUCTION}

Path planning is a prevalent problem that exists, for example, in traffic control to optimize traffic flow patterns, but also in robotics for finding the best way to a target or to plan an arm-hand trajectory when performing manipulation.  Even protein-folding can be formulated as a path planning problem \cite{Amato2002} and additional relevant applications exist in other fields. In general, path planning is defined as the problem of finding a temporal sequence of valid states from an initial to a final state given some constraints \cite{Latombe2012}. 

In this contribution we are addressing path planning for multiple start- and end-points  (i.e., multi-source multi-target) in large environments, like city maps. This problem relates to the single source shortest path (SSSP) problem. The task of SSSP is to find shortest paths in a graph between a vertex (node) and all other vertices such that the sum of the edges' weights is minimised. Classical approaches to solve SSSP are Breadth First Search algorithm (BFS) \cite{Moore1959}, Dijkstra's algorithm \cite{Dijkstra1959}, and the Bellman-Ford algorithm \cite{Bellman1958,Ford1956}. BFS is suitable for unweighted graphs, i.e., is a uniform cost search and the obtained solution is optimal with respect to the number of nodes to travel from the source node to all other nodes. Dijsktra's algorithm finds shortest paths in weighted graphs with positive weights, whereas the Bellman-Ford algorithm can also deal with graphs with negative weights. However, it is slower than Dijsktra's algorithm. Another algorithm which finds shortest paths between all pairs of nodes in a graph is the Floyd–Warshall algorithm \cite{Floyd1962}, however, running Dijkstra's algorithm for each node is a better choice when considering sparse graphs. The advantage of all these methods is that they do not require learning and are parameter free methods, however, they can be computationally expensive.

Recently, an new approach, which also does not require learning has been proposed by \cite{Farias2019}. This method utilises GPU OpenGL shaders and is based on cone rasterization from sources and obstacle vertices. It can generate optimal path maps for multiple sources and outperforms other GPU based approaches \cite{Luo2010,Merrill2012,Wynters2013,Kapadia2013,Garcia2014}. However, as we will show later, this method is not very suitable in several cases, because it leads to relative long computational times. 

Another class of algorithms is based on artificial neural networks ranging from bio-inspired approaches to deep learning methods. In the bio-inspired approaches \cite{Glasius1995,Glasius1996,Bin2004,Yang2001,Ni2017,Rueckert2016}, the environment is represented by a network with inhibitory (=obstacles) and excitatory (=free spaces) neurons arranged on a grid. Here, activity is propagated from the source neuron to the closest neurons and this procedure is repeated for many iterations until the activity is spread out across the whole network. Later, shortest paths can be found by following activity gradients. In principle, these networks are similar to the wavefront propagation algorithm \cite{Choset2005}, which is a special case of the BFS algorithm. Although most of these these approaches (except \cite{Rueckert2016}) do not require network training, they are not parameter free. Also, as we will show later, the disadvantage of such algorithms is that the activity decreases exponentially \cite{Yang2001} and large mazes (e.g., larger than $500 \times 500$) can not be solved due to numerical precision problems.

Recently, deep learning approaches utilising deep multi-layer perceptrons (DMLP, \cite{Qureshi2018}), fully convolutional networks \cite{Perez2018,Ariki2019,Kulvicius2020}, long short-term memory (LSTM) networks \cite{Bency2019}, and deep reinforcement learning approaches \cite{Tai2017,Panov2018} have been proposed for solving path finding problem, too. Most of these approaches (except \cite{Ariki2019} and \cite{Kulvicius2020}) deal with single path planning and/or consider relatively small environments (below $250 \times 250$). Moreover, these approaches require relatively large data sets (e.g., thousands of samples) as well as training to optimise network parameters.

In this paper we present a novel deep network, which consists of only max pooling layers (oMAP). The proposed method generates activity maps for single as well as for multiple sources to single but also multiple targets. It does not require training data such that learning is not necessary and, when following the activity gradient shortest paths are found. Furthermore, this approach can process very large environments on standard GPUs in very short time\footnote{A preprint of this paper has been uploaded to \url{https://arxiv.org}}.
\begin{figure}[!t]
\begin{center}
\includegraphics[width=1.0\linewidth]{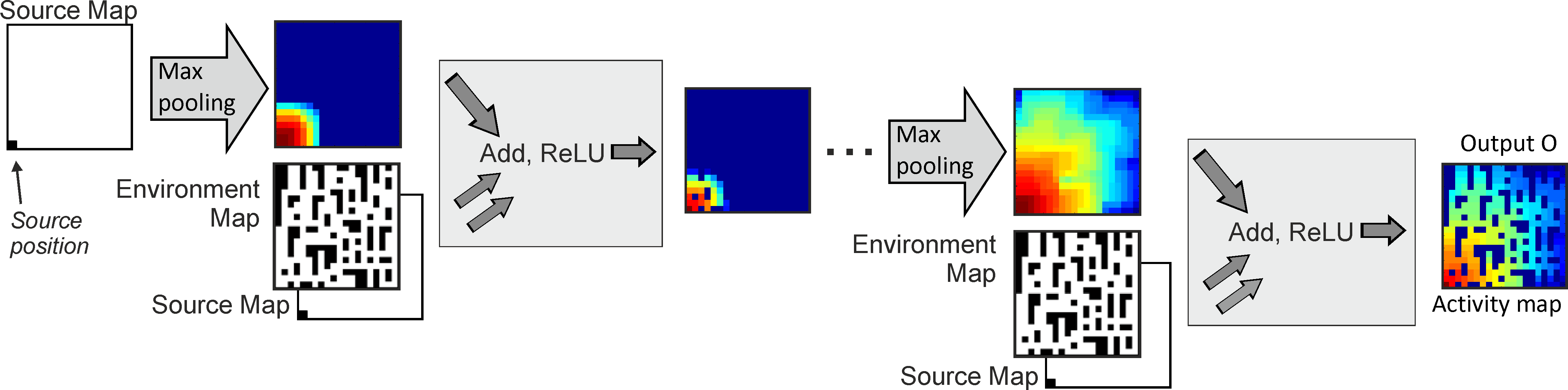}
\caption{\label{net} Network architecture and algorithmic process. The network consists of many stacked identical max pooling layers with one filter of size $3 \times 3$, stride $1 \times 1$ and zero padding (no sub-sampling). Here, for graphical reasons, a wider filter is shown. The network receives a source map and an environment map (with obstacles, black) as inputs. The algorithm consists only of repeated max pooling, adding, and rectified-linear (ReLU) operations as shown in the figure.}
\end{center}
\end{figure}

\section*{\flushleft METHODS}

\subsection*{\textbf{Input}}
The oMAP algorithm uses two binary images of size $m \times n$, an environment map $I_e$ and a source map $I_s$, with $s\ge 1$ sources, as an input.

For the source map, we set $I_s(i,j)=1$ at all source locations, otherwise we set $I_s(i,j)=0$.

The environment map $I_e$ represents an obstacle map where we set $I_e(i,j)=0$ if a grid cell is free (no obstacle) and $I_e(i,j)= -maxint$ if grid cell $(i,j)$ contains an obstacle. The choice of using here the numerically most negative integer ($-maxint$) is motivated by the algorithm for generating the activity map, described next. 

\subsection*{\textbf{Activity Map Generation}}
The oMAP network consists of $L$ identical max pooling layers $l_i$ (see Fig.~\ref{net}) with only one type of filter with size $3 \times 3$. We specifically use such a filter size in order to pass activity only to the nearest grid cells (similar to the wavefront expansion algorithm \cite{Choset2005}). Otherwise, in case of larger filters, activity could propagate also to grid cells, which are separated by obstacles.

\begin{figure}[!t]
\begin{center}
\includegraphics[width=0.99\linewidth]{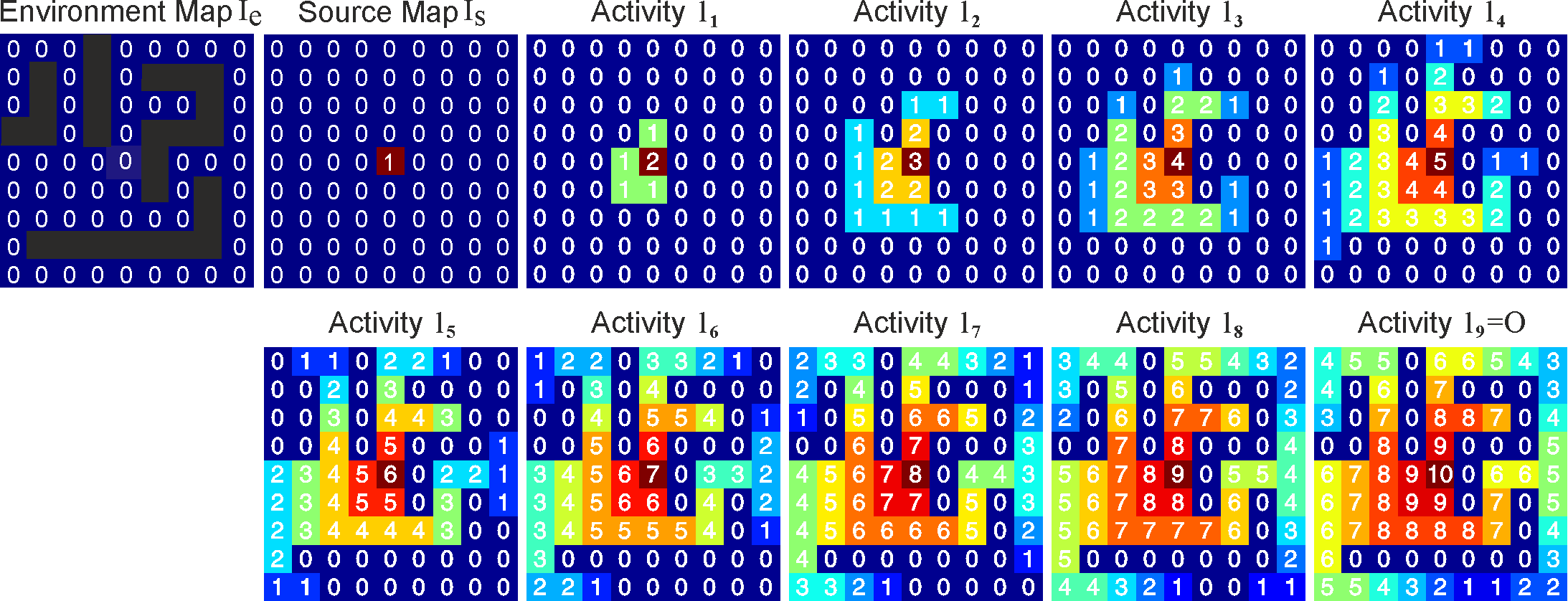}
\caption{\label{map_ex} Illustration of the generation of an activity map using oMAP. Black grid cells in the Environment Map stand for cell values of $-maxint$ and represent obstacles. Brown cell with value 1 in the Source Map denotes the source location. Activity at each layer after Add/ReLU operation (see Fig.~\ref{net}) is shown. }
\end{center}
\end{figure}

We start with the source map $I_s$ and perform max pooling. Then we sum the resulting map with input maps $I_e$ plus $I_s$ into one intermediate-layer map. Note that the largest resulting grid cell value $v$ after this operation will be $v=n+1$ (see example in Figure~\ref{map_ex}), where $n$ is the index of the current layer $l_n$. All source cells will obtain this value, hence $v_{source}=n+1$. Obstacle cells, on the other hand, obtain values, which remain negative and follow $v_{obst}<-maxint+n+1 < 0$.

Then we pass the resulting map through the standard rectified linear transfer function (ReLU). Because $v_{obst}< 0$ all grid values at obstacle locations remain at zero.

This is repeated for all layers until the last layer, which produces the final activity map $O$ of size $m \times n$ as an output. 

Note that (different from, e.g., convolutional nets) we do not have any tunable weights, thus, no training is required.

A graphical visualisation of the process of activity map generation using oMAP is shown in Fig.~\ref{map_ex}. Here we used a maze of size $9 \times 9$ and placed one source in the middle. We show the obstacle map (black grid cells denote obstacles), the input activity (source map) and the activity of each max pooling layer after Add/ReLU operation. As we can see, after the first layer (see output $l_1$) activity is propagated from the source only to its neighbouring grid cells (except obstacles) which obtain values of $1$, while the activity at the source cell is increased by one to a value of $2$. From layer to layer, activity in the network grows and propagates to grid cells increasingly distant from the source. In this particular example, map generation is complete after nine layers. 

\subsubsection*{\textbf{Determining the number of layers $L$ and algorithmic complexity}}
oMAP does not have any tunable variables and also the number of layers $L$, which is the only existing free parameter, can be unequivocally determined. It is identical to the maximal path length in an environment, which depends on the location of the source(s), the size of the environment and the distribution of obstacles.
In general, however, the longest path is {\it a priori} unknown, but the structure of the oMAP algorithm allows determining $L$ during run-time. The algorithm can be run recursively adding layer after layer until the activity map does not contain any zero-values anymore (except at the obstacles). Thus, $L$ can be set using this procedure\footnote{Note, in the Appendix we will show that recursive running of oMAP is very slow due to a per-iteration required CPU-GPU handshake. Thus, for practical purposes, we did not run oMAP in recursive mode and determined $L$ instead by prima-vista estimating path complexity.}.

The complexity of our algorithm is $\mathcal{O}(N \times L)$ where $N = m \times n$ is the number of grid cells in the map (corresponds to the number of $max$ operations per layer) and $L$ is the number of layers.

\subsection*{\textbf{Path Reconstruction}}

A path from any given target location to the source (or closest source) can be found from the generated activity map $O$ by following the activity gradient; i.e., we start from the chosen target location and select a neighbouring grid cell out of its eight neighbors with maximum value and repeat this until reaching the source. Note that there can be cases of more than one neighboring cell with maximal value. In such a case, we chose the next cell randomly and we refer to this method as \textit{simple path reconstruction}. Note that, due to the single step forward propagation by the max pooling method described above, it will not matter, which cell to choose, because following any of the resulting gradients will render the same number of steps back to the source. 

Using map $l_9$ from Fig.~\ref{map_ex}, it is easy to see that this method can create all possible paths from any target back to the source in the middle.

Note that this method is similar to the wavefront expansion algorithm \cite{Choset2005}, which is a special case of  Breadth First Search (BFS, \cite{Moore1959}) and, thus, always renders optimal paths with respect to number of steps.

However, paths obtained by simple path reconstruction will be not necessarily optimal with respect to Euclidean distance. This is due to the fact, that all transitions (horizontal, vertical, diagonal) are weighted equally amounting to a uniform-cost search. To improve on this, we propose the {\it Euclidean path reconstruction} method, described next.

We first find a path, similar to above, by choosing a neighboring cell with maximum value, but now we only consider horizontal and vertical neighboring cells ('Manhattan' transition). Thus, given the current path position $\{P_x(t)=i,P_y(t)=j\}$ the next step of the path is defined by
\begin{eqnarray}
    \{P_x(t+1),P_y(t+1)\} = \arg\max_{i,j} \{O(i,j-1),\\
    \nonumber O(i,j+1),O(i-1,j),O(i+1,j)\}.
\end{eqnarray}

We stop this path search as soon as the (nearest) source is reached. Then we straighten the path by removing intermediate points  $P_{x,y}(t)$ if $||P_{x,y}(t-1)-P_{x,y}(t+1)|| = \sqrt{2}$, $t=2 \dots k-1$, where $k$ is the number of points in the path. Note that complexity of the path reconstruction procedure is $\mathcal{O}(k)$. The obtained path is now shortest with respect to the Euclidean distance. 

oMAP was implemented using Tensorflow and Keras API\footnote{The data and source code will be published online after acceptance.}. We used a PC with Intel Xeon Silver 4114 CPU (2.2GHz) and NVIDIA GTX 1080 Ti or NVIDIA Titan V GPU. 

\begin{figure}[!t]
\begin{center}
\includegraphics[width=1.0\linewidth]{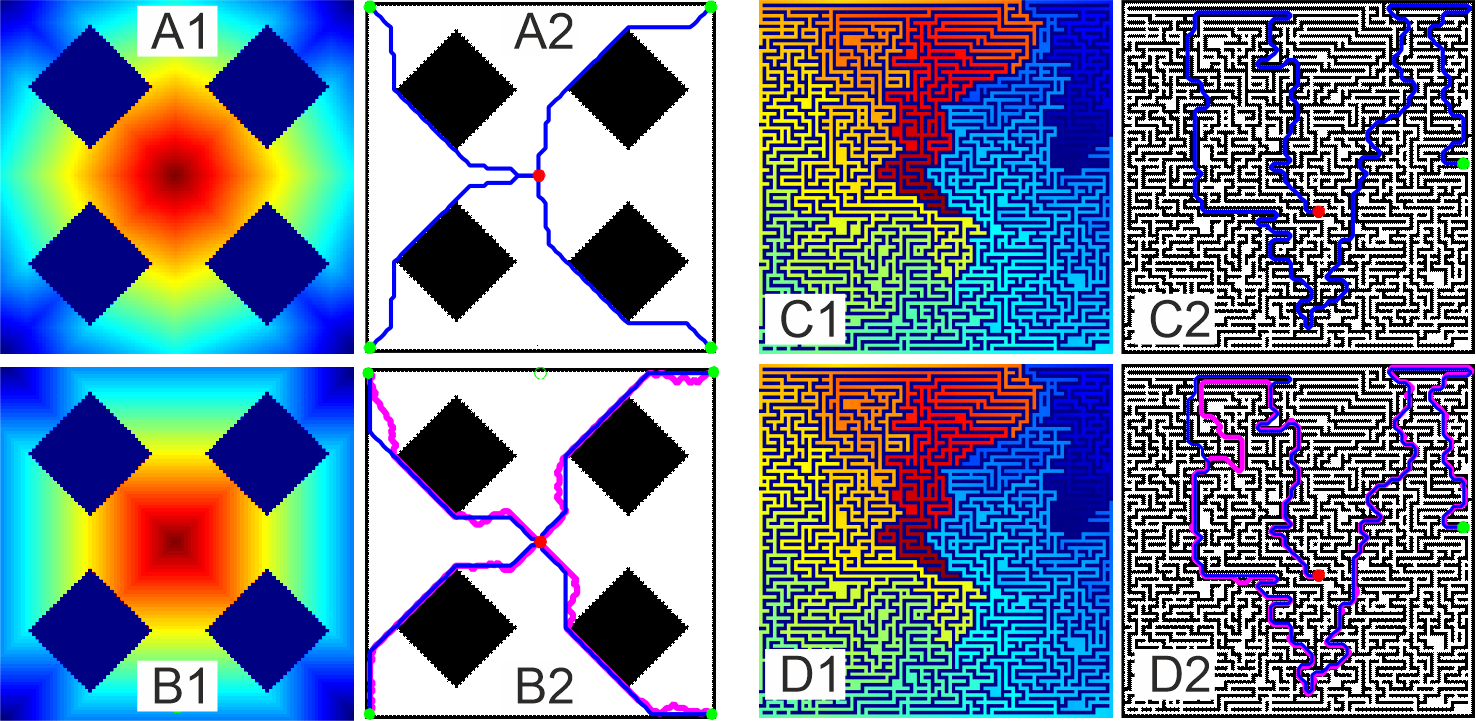}
\caption{\label{dijkstra_vs_max} \textbf{A, C)} Dijkstra's algorithm versus \textbf{B, D)} oMAP. Activity maps (index 1)  and reconstructed trajectories (index 2) are shown. Grid size is $100 \times 100$ in case A, B, and $101 \times 101$ in case C, D. For oMAP we used 100 layers for B and 450 layers for D. Blue trajectories represent optimal (shortest) paths following Euclidean path reconstruction and magenta trajectories represent non-optimal paths following simple path reconstruction. Green and red dots represent start- and end-points, respectively.}
\end{center}
\end{figure}

\section*{\flushleft RESULTS}

\subsection*{\textbf{Generation of Activity Maps and Path Reconstruction for oMAP and other Algorithms}}

We compared our approach to several above mentioned algorithms, which do not require learning, i.e., Dijkstra's algorithm \cite{Dijkstra1959}, a biologically inspired neural network (shunting model) \cite{Yang2001} and a state-of-the-art algorithm based on OpenGL shaders \cite{Farias2019}. We assessed path optimality (shortest paths with respect to the Euclidean distance) as well as run-time. For the run-time comparison against OpenGL shaders we used the benchmark maps as in \cite{Farias2019}.

\subsubsection*{\textbf{oMAP versus Dijkstra}}
First, we show qualitative results on the generation of avtivity maps with single and multiple sources using oMAP compared to Dijkstra's algorithm. Examples of activity map generation and path reconstruction are shown in Fig~\ref{dijkstra_vs_max} in panels A, C for Dijkstra and in panels B, D for oMAP. Here we used two artificial environments, a relatively simple map with four obstacles (reproduced from \cite{Farias2019}, panels A and B) and a complicated maze map (generated automatically, panels C and D). We show normalised activity maps between zero (dark blue) and maximum (dark red). We obtain a circular pattern when using Dijkstra (see panel A1) and a square pattern with oMAP (B1). This is due to the fact that Dijkstra uses a non-uniform cost (horizontal/vertical moves have a cost of $1$ and diagonal moves have a cost of $\sqrt{2}$), while for oMAP we use uniform cost (all moves have a cost of $1$).

In case of Dijkstra's algorithm (panel A2) paths were reconstructed from the \textit{visited nodes list} \cite{Dijkstra1959} and are of optimal Euclidean length, in spite of their 'wiggly' appearance. On the other hand, maps generated by oMAP combined with simple path reconstruction (magenta paths in B2, D2) will lead to optimal paths with respect to the number of steps, but these paths are not optimal with respect to Euclidean distance. Euclidean path reconstruction (blue paths) solves this issue. Note that for this, one follows first (before straightening) the gradient only along Manhattan transitions. This will usually not lead to the same grid cell selection as for simple path reconstruction. Hence blue and magenta paths are independent of each other. This can be seen when comparing the reconstructed paths (see, for example, panel B2) from simple path reconstruction (magenta) with those from Euclidean path reconstruction (blue). The latter renders optimal paths, which are straighter than the ones from Dijkstra.  Also for complex mazes, Euclidean reconstruction renders optimal paths (C, D), in this case identical to Dijkstra. 

\begin{figure}[!hbt]
\begin{center}
\includegraphics[width=1.0\linewidth]{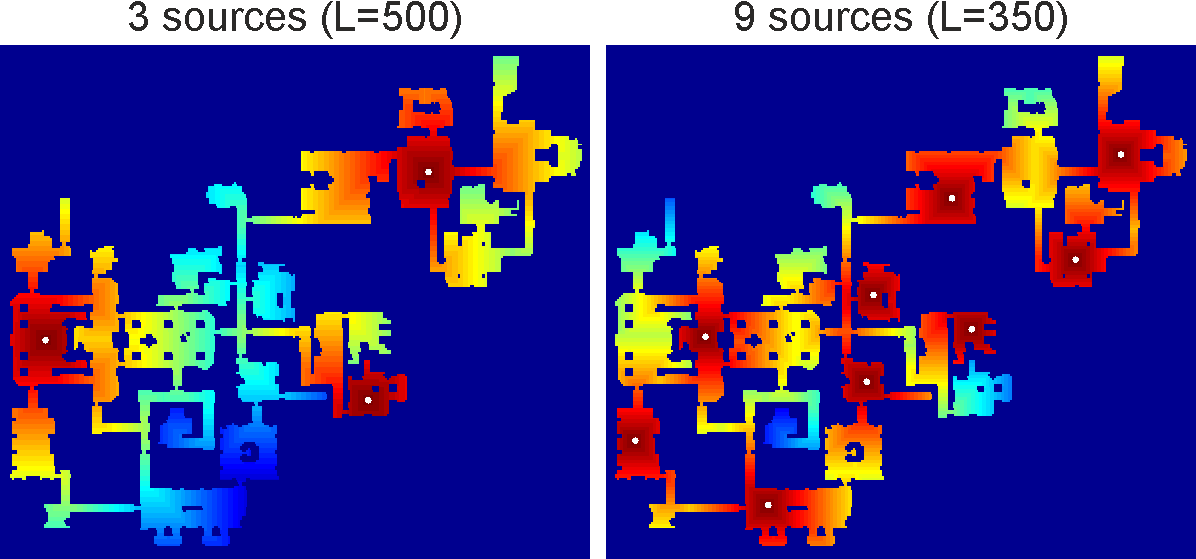}
\caption{\label{multi_source} Examples of activity map generation using three (left) and nine (right) sources (white dots). Here we used a map from Moving AI benchmark \cite{MovingAI} (map resolution $767 \times 881$). }
\end{center}
\end{figure}

An example of activity map generation by oMAP using three and nine sources is shown in Fig.~\ref{multi_source}. For this, we used a map from the Moving AI benchmark \cite{MovingAI} of resolution $767 \times 881$. We used 500 layers and 350 layers in case of three and nine sources, respectively. Results demonstrate that less layers are needed if the number of sources increases, since activity propagates from all sources at the same time, which, as a consequence, fills the complete map sooner.

\subsubsection*{\textbf{oMAP versus Shunting Model}}

\begin{figure}[!hbt]
\begin{center}
\includegraphics[width=1.0\linewidth]{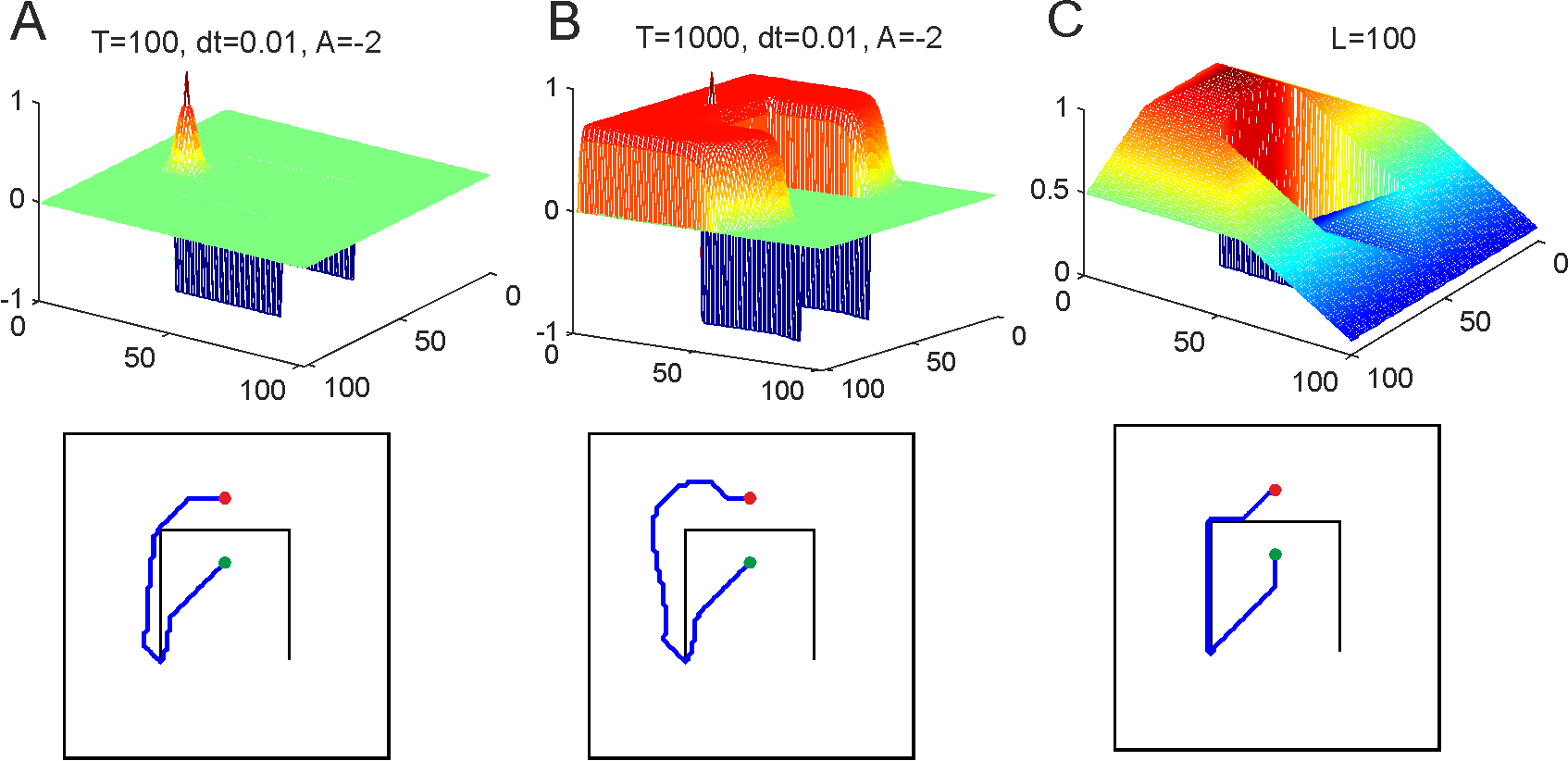}
\caption{\label{bio_vs_max} \textbf{A, B)} Biologically inspired neural network \cite{Yang2001} versus \textbf{C)} oMAP with 100 layers. Grid size is $100 \times 100$. Greed and red dots represent start- and end-points, respectively. A 3D plot has been used to more clearly show the structure of the gradients.}
\end{center}
\end{figure}

Conceptually, our approach, as already discussed in the Introduction section, is similar to the path finding method using a biologically inspired neural network (shunting model, \cite{Yang2001}). Thus, we also compare our approach to this method. Results obtained on a map with a u-shape obstacle (a common benchmark for the evaluation of path finding methods; reproduced from \cite{Yang2001}) is shown in Fig.~\ref{bio_vs_max}. The central disadvantage of the shunting model is that large environments (e.g. above $500 \times 500$) cannot be addressed. Either neuronal activity drops very quickly when moving away from the source (see panel A) and soon reaches 'numerical-zero'. Or, for quite long run-times, activity can indeed be propagated to more distant locations but this easily leads to activity plateaus due to the nature of the model (see panel B).
The authors comment on their model parameters \cite{Yang2001} but even after extended search in the parameters space, we were not able to arrive at a shunting model that could solve environments above $500 \times 500$. Thus, the shunting model is only applicable for relatively small environments. Moreover, the authors of that study state \cite{Yang2001} that their method generates optimal paths, which is not always the case. The paths in Fig.~\ref{bio_vs_max}~A, B are non-optimal, due to the fact that activity of each neuron is computed as a weighted average activity of its nearest neighboring neurons, which may lead to sub-optimal paths in environments with obstacles. 

In contrast to that, activity decreases linearly from the source when using oMAP, which allows generating activity maps for extremely large environments, using large enough $L$, and numerical precision problems do not exist, because we operate with integer numbers that grow maximally to the longest path length. As a consequence, the resulting paths are always optimal (Fig.~\ref{bio_vs_max}~C).

\subsubsection*{\textbf{oMAP versus OpenGL Shaders}}
\begin{figure}[!t]
\begin{center}
\includegraphics[width=1.0\linewidth]{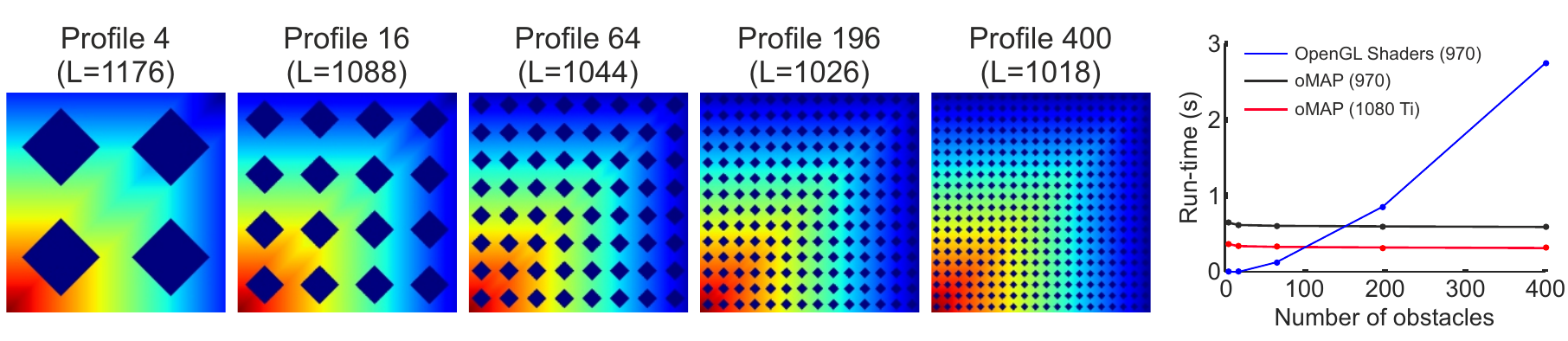}
\caption{\label{time_OGLS} Run-time comparison between OpenGL shaders and oMAP for different types of hardware (NVIDIA GTX 970 vs. NVIDIA GTX 1080 Ti). Left: Activity maps generated using oMAP for the profile maps of size $1,000 \times 1,000$ reproduced from \cite{Farias2019} with 4, 16, 64, 196, and 400 obstacles. The number of layers for oMAP is given in parenthesis. The source was at the bottom left corner in all cases. Right: run-time comparison for these profiles. Data for the blue curve were taken from \cite{Farias2019}.}
\end{center}
\end{figure}

Possibly the most powerful, currently existing method that uses a wavefront propagation algorithm is described by \cite{Farias2019} and employs OpenGL shaders in a GPU implementation. This method will always produce optimal paths. Therefore, we chose to compare it to oMAP according to the run time of both algorithms (Fig.~\ref{time_OGLS}). Note, however, that such across-implementation comparisons have to be taken with a grain of salt, because we cannot know how efficient the foreign implementation was.

For this, we used maps of size $1,000 \times 1,000$ with increasing number of obstacles as in \cite{Farias2019} (see obstacle configurations on the oMAP activity maps in Fig.~\ref{time_OGLS}, left). Note that in our case we set the source in the bottom-left corner, which is the worst case with respect to  computer time. Results demonstrate that the run-time of the OpenGL shaders method depends on the number of obstacles in the scene and this method slows down non-linearly as the number of obstacles increases. By contrast, our method uses almost constant time and is faster than \cite{Farias2019} as soon as the number of obstacles increases. The slight run-time decrease observed for oMAP is due to the fact that the largest source-to-target distance in the map is decreasing as obstacles become smaller, and fewer layers are needed to create the map. Note that for oMAP run-time scales down by a factor of two if the source were located in the middle (best case). The different curves show the performance for different GPUs. Farias and Kallmann \cite{Farias2019} had used the older NVIDIA GTX 970 (blue). The performance of oMAP was also determined according to this hardware (black), but also when using the faster NVIDIA GTX 1080 Ti (red), which had been also used for all other experiments.

\subsection*{\textbf{Multi-source Multi-target Path Planning}}

\begin{figure}[!hbt]
\begin{center}
\includegraphics[width=0.75\linewidth]{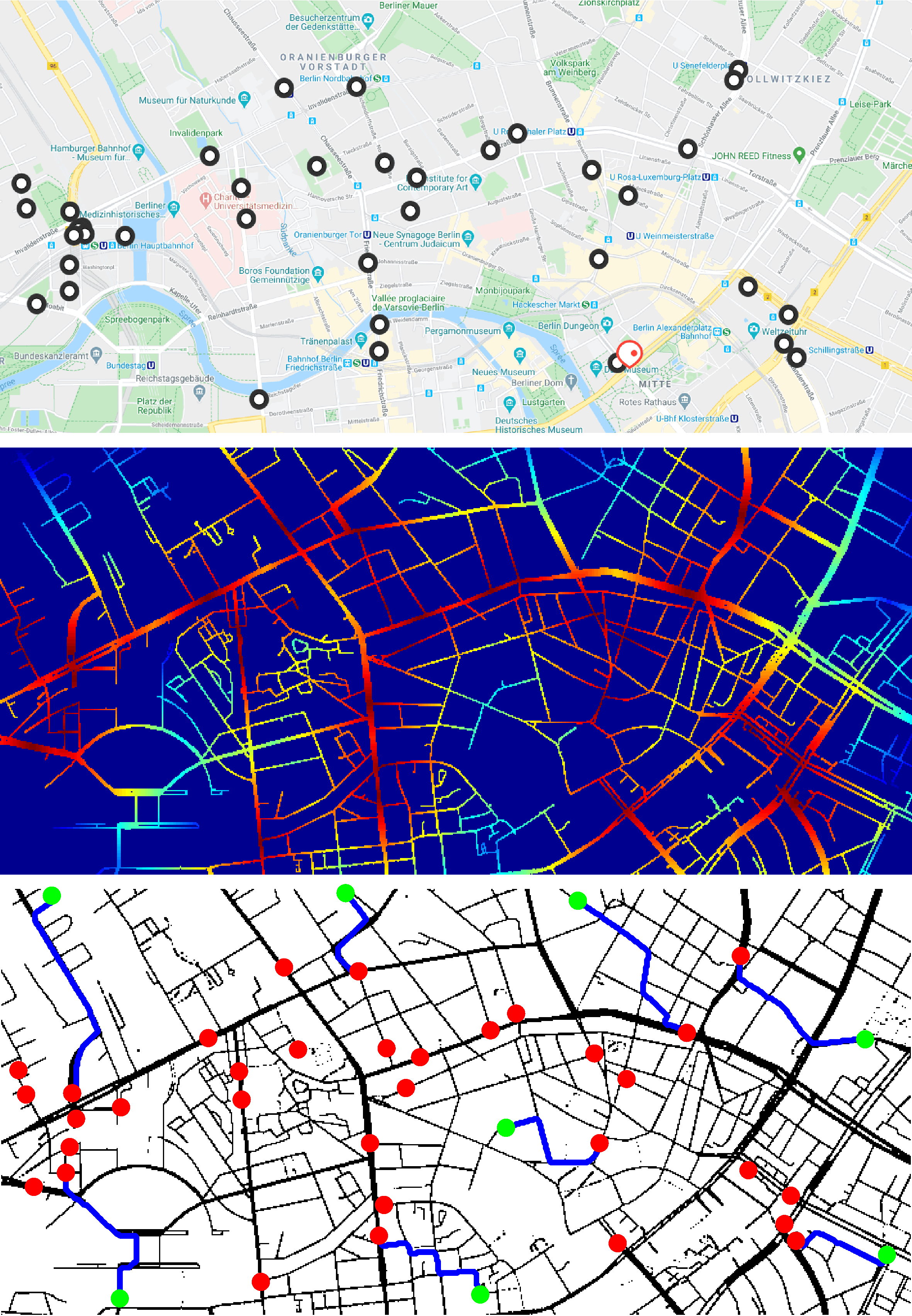}
\caption{\label{res_multi} Multi-source - multi target path finding problem in a real taxi scenario. Top: Section of a Berlin city map ($332 \times 709$) with disks showing taxi positions obtained on March, 03, 2020 at 2:19 pm, which is a screenshot from the on-line application available at \textit{https://www.taxi.de}.  Middle: Multi-source activity map using oMAP (160 layers). Bottom: Shortest paths (blue lines) from customers (green dots) to taxis (red dots) are shown. Customer positions were defined manually.}
\end{center}
\end{figure}

Finally, we applied our method to a  multi-source - multi-target task testing our network on a real taxi scenario in Berlin, where there are multiple taxi cabs and multiple customers. The task was to find the closest taxi cab for each customer. To be realistic as possible, we obtained taxi cab positions in the area of Berlin close to the train station at a specific moment in time using an on-line taxi application provided by \url{https://www.taxi.de}. The streets were extracted automatically using our own written program. Positions of the taxi cabs were set as sources (in total 33; taxi cabs which were very close to each other were marked with one source) and an activity map was generated using oMAP. Finally, eight customer positions were defined manually and the optimal paths were reconstructed from the activity map for each customer as described above. Results are shown in Fig.~\ref{res_multi}. In the top panel we show taxi locations in Berlin close to the main train station from an internet-taxi-app at one given point in time. In the middle and the bottom panel we show the resulting activity map for 33 sources and the reconstructed shortest paths for eight customers to the closest taxi cabs, respectively. We used a network with 160 layers and it took only 20 $ms$ on NVIDIA GTX 1080 Ti to generate the activity map (resolution $332 \times 709$). The reconstructed paths (blue trajectories) show that the closest taxi cabs and the shortest paths were found for all eight customers. 

\subsection*{\textbf{General Run-time Evaluation, Limitations and Practical Considerations}}

\subsubsection*{\textbf{Run-time}} Run-time evaluations evidently depend on the hardware used. Still, it is of interest to document where we stand given currently existing state of the art hardware. In the following we will therefore show that oMAP achieves remarkable performance when tested on an NVIDIA GTX 1080 Ti GPU.  We define the linear grid size as $n$ and consider here square grids with $n^2$ grid cells ("nodes"). We used empty maps without obstacles because for oMAP the number of operations does not depend on the number of obstacles. 

\begin{figure}[!hbt]
\begin{center}
\includegraphics[width=1.0\linewidth]{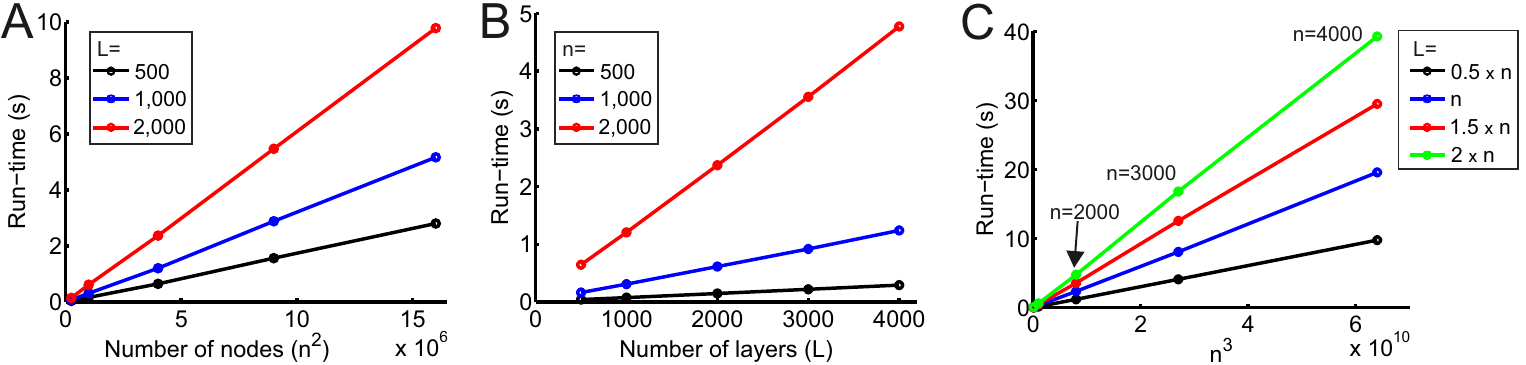}
\caption{\label{time_pro} Run-time evaluation of oMAP. \textbf{A)} Run-time versus different number of nodes $n^2$ (linear grid size: $n$ = [500; 1,000; 2,000; 3,000; 4,000] for a fixed number of layers $L$. \textbf{B)} Run-time versus different number of layers ($L$=[500; 1,000; 2,000; 3,000; 4,000]) for a fixed linear grid size $n$. \textbf{C)} Run-time versus $n^3$ ($n$ = [500; 1,000; 2,000; 3,000; 4,000]) and different number of layers ($L$=[0.5$\times n$, $n$, 1.5$\times n$, 2$\times n$]). Slopes of the lines (from black to green) are: $1.5265\times10^{-10}$, $3.0593\times10^{-10}$, $4.6202\times10^{-10}$, $6.1493\times10^{-10}$.}
\end{center}
\end{figure}

Panel A in Fig.~\ref{time_pro} shows that run-time increases linearly against the number of nodes $n^2$ when keeping the number of layers constant. Similarly, linear growth is also observed when keeping the grid size $n$ constant and increasing $L$ (Fig.~\ref{time_pro}~B). 

From above it is clear that more layers are needed when the grid gets bigger, because this leads to longer paths for which the network has to be increased. The shortest possible longest-path length is $0.5n$ (empty square grid, source in the middle\footnote{We are counting steps, where diagonal moves count the same as horizontal/vertical ones.}). In panel C we, thus, consider the number of nodes $n^2$ together with a changeable number of layers $L$ and we let $L$ depend on the grid size for approximating the fact that in larger grids paths are longer. As expected from panels A and B these curves now linearly follow $n^3$ with slopes that increase for increasing $L$ also in a linear manner (for slope values see figure caption). From all this, an approximate equation for estimating the run-time for different grid sizes and different number of layers can be derived as $t_r(n,L)\approx 3.0751 \times 10^{-10} L \, n^2$.

\begin{figure}[!hbt]
\begin{center}
\includegraphics[width=1.0\linewidth]{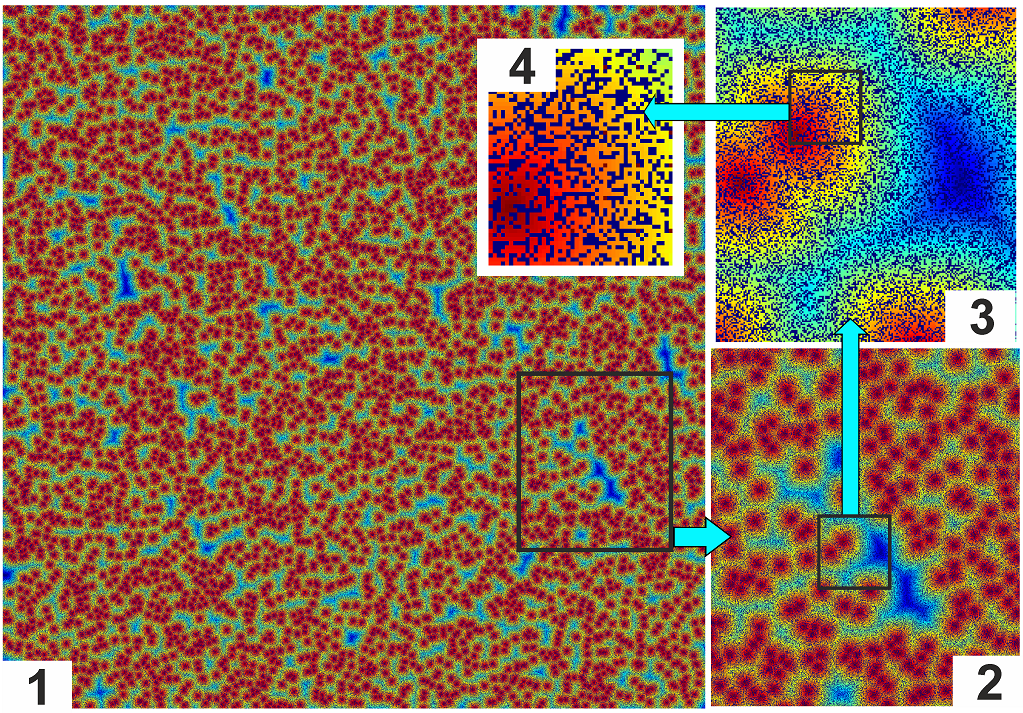}
\caption{\label{big} Example of a huge grid with $n=26,000$ corresponding to 676,000,000 nodes. Panel 1 shows the full grid with 5,000 sources and panels 2 to 4 show magnifications to make the obstacles visible (panel 4). Colors encode activity as in Figure~\ref{map_ex}, blue=small and red=large values.}
\end{center}
\end{figure}

\subsubsection*{\textbf{Limitations and Practical Considerations}} The above run-time estimate holds on an NVIDIA GTX 1080 Ti GPU as long as oMAP runs essentially in forward mode without (too many) iterations $i$. Note, however, that the overhead of having to pass information iteratively back to the start of a new batch of layers will remain tiny if this happens just a few times.

As stated above, the number of required layers $L$ depends on the longest path. The theoretically existing longest path in any $n\times m$ grid is given as $\max(n,m) \times int((\min(n,m)+1)/2)  + int(\min(n,m)/2)$, which is a path that meanders back and forth between two interleaved comb-like obstacle rows. In a square grid, this number can be approximated by $\frac{n^2}{2}$ and this number would have to be matched by $L$. From our experiments with real maps, we found, however, that this is a highly unrealistic situation and usually $1.5n<L<2n$ suffices. 

For example, on this architecture, we could run oMAP in one forward pass for a remarkable linear grid size of $n_{max}=26,000$, equivalent to 676,000,000 nodes (see panel 1 in Fig.~\ref{big}) with maximal layer number of $L_{max}=3,450$ (on an NVIDIA GTX 1080 Ti GPU). This would correspond to a panel of $3.25\times3.25$ tiles when using 64 Megapixel images as individual "maps" with every pixel a grid cell.  Run-time for this case was $692.3~s$. Interestingly, for this system $L_{max}$ does not depend on the grid-size. 

Note that the number of required layers $L$ will decrease, and so does the run-time, as soon as more than one source exists. Figure~\ref{big} shows an example of a grid with $n=26,000$ and $5,000$ sources. This required $650$ layers and took only $147.4~s$ to run.

Some more examples that show the power and the limitation of oMAP are: if we assume $L=2 \, n$, then we can run systems with $n=1725$ in one forward pass in  $\approx 3.1~s$. Under the same assumption ($L=2 \, n$) we would need $L=52,000$ for the maximal possible grid with $n=26,000$. This would require $i=15$ full iterations and a few more layers for iteration 16, resulting in a run time of $t_r\approx 10,800~s$ (3 hours).

\section*{\flushleft CONCLUSION}
We have presented a deep network for path finding in grid-like environments that does not need learning and runs very fast even for  large environments with complex paths. It outperforms competing network approaches by a large margin and is easy to implement on standard GPUs, because of its simple structure. There are no free parameters except the number of layers $L$ for which, however, efficient approximations exist.

For example, in a square grid, the maximal path-length (worst case) can be approximated by $\frac{n^2}{2}$, which would have to be matched by $L$. However, realistically we found that $n<L<2 \, n$ was most often sufficient. The extreme case with highly meandering paths in Fig. ~\ref{dijkstra_vs_max}~D, which in practical, map-like situations is unlikely to exist, needed $L\approx 4.5 \, n$.

Furthermore, $L$ decreases, when more sources are introduced. Thus, the proposed oMAP approach has, in particular, high potential for multi-source multi-target applications in large environments.

\section*{\flushleft APPENDIX}

\subsection*{\textbf{Run-time Evaluation for Iterative versus One-shot Implementation}}

We also compared the run-time of two different implementations of oMAP on two different GPUs, NVIDIA  GTX 1080 Ti and NVIDIA Titan V. We compared two extreme cases, where in one case we ran oMAP with 2,000 max pooling layers for one iteration (one forward-pass) and in the other case we ran a single-layer oMAP network for 2,000 iterations recursively, i.e., the output was used as the next input. Results are shown in Table~\ref{gpu_comp}  where we can see that the former architecture is two orders of magnitude faster as compared to the latter, due to processing and passing data between the CPU and GPU, as compared to the case where all operations are processed solely on the GPU. Results also demonstrate that a significantly larger speed-up can be achieved with the multi-layer architecture as compared to the single-layer architecture when the faster GPU is used (2.72 times versus 1.07).    

\begin{table}[!hbt]
\begin{center}
\caption{\label{gpu_comp} Run-time comparison between NVIDIA GTX 1080 Ti and NVIDIA Titan V for oMAP with a recursive versus a linear implementation. Grid size of $2,000 \times 2,000$ was used in all cases.}
\begin{tabular}{|l|c|c|}
\hline
\begin{tabular}[c]{@{}l@{}}Number of layers /\\ Number of iterations\end{tabular} & \textbf{2,000 / 1} & \textbf{1 / 2,000}\\ \hline
\begin{tabular}[c]{@{}l@{}}Run-time (s) on\\ \textbf{NVIDIA  GTX 1080 Ti}\end{tabular} & 2.370 & 117.314\\ \hline
\begin{tabular}[c]{@{}l@{}}Runt-time (s) on\\ \textbf{NVIDIA Titan V}\end{tabular} & 0.871 & 109.299\\ \hline
\end{tabular}
\end{center}
\end{table}




\bibliographystyle{plain}

\end{document}